\documentclass[journal]{IEEEtran}

\ifCLASSINFOpdf
\else
\fi

\ifCLASSOPTIONcompsoc
    \usepackage[caption=false, font=normalsize, labelfont=sf, textfont=sf]{subfig}
\else
	\usepackage[caption=false, font=footnotesize]{subfig}
\fi

\usepackage{cite}
\usepackage[version=4]{mhchem}
\usepackage{graphicx}
\usepackage{amsmath,amssymb,amsfonts}
\usepackage{color,soul}
\usepackage[table,xcdraw]{xcolor}
\usepackage{romannum}
\usepackage{graphicx}
\usepackage{commath}

\graphicspath{{figures/}}

\begin{document}

\title{A Fully Memristive Spiking Neural Network with Unsupervised Learning}

\author{\vspace{-10pt}
Peng Zhou,~\IEEEmembership{Student Member,~IEEE,}
        Dong-Uk Choi,
        Jason~K.~Eshraghian,~\IEEEmembership{Member,~IEEE,}\\
        and Sung-Mo Kang,~\IEEEmembership{Life~Fellow,~IEEE}
        
\thanks{P. Zhou, D. Choi, and S. M. Kang are with the Department of Electrical and Computer Engineering, UC, Santa Cruz, CA, USA.}%
\thanks{J. K. Eshraghian is with the Dept. of Electrical Engineering and Computer Science, University of Michigan, MI, USA and the Dept. of Computer Science and Software Engineering, University of Western Australia, Australia.}%
}

\maketitle
\vspace{-10pt}
\begin{abstract}
We present a fully memristive spiking neural network (MSNN) consisting of physically-realizable memristive neurons and memristive synapses to implement an unsupervised Spike Timing Dependent Plasticity (STDP) learning rule. The system is fully memristive in that both neuronal \textit{and} synaptic dynamics can be realized by using memristors. The neuron is implemented using the SPICE-level  memristive integrate-and-fire (MIF) model, which consists of a minimal number of circuit elements necessary to achieve distinct depolarization, hyperpolarization, and repolarization voltage waveforms. The proposed MSNN uniquely implements STDP learning by using cumulative weight changes in memristive synapses from the voltage waveform changes across the synapses, which arise from the presynaptic and postsynaptic spiking voltage signals during the training process. 
Two types of MSNN architectures are investigated: 1) a biologically plausible memory retrieval system, and 2) a multi-class classification system. Our circuit simulation results verify the MSNN’s unsupervised learning efficacy by replicating biological memory retrieval mechanisms, and achieving 97.5$\%$ accuracy in a 4-pattern recognition problem in a large scale discriminative MSNN.

\end{abstract}

%
\IEEEpeerreviewmaketitle

\vspace{-4pt}
\section{Introduction}
\IEEEPARstart{N}{euromorphic} computing is guided by the rich neural dynamics present in the brain, the highly parallelized nature of neural computation, and the sparse encoding of data as spikes, in pursuit of optimizing memory and computation for energy efficiency. The pervasive von Neumann architecture disaggregates memory and computation which leaves much room for improvement for threads with a deterministic set of instructions. This deficiency has spurred the development of a variety of neuromorphic computing systems \cite{davies2018loihi, furber2014spinnaker,benjamin2014neurogrid, merolla2014million, schemmel2010wafer, moradi2017scalable, camunas2019neuromorphic, furber2016large, orchard2021efficient}, which integrate spiking neurons and simplified synaptic models onto a silicon substrate. In almost all instances, synaptic weights are stored in random access memory (RAM), thus moving memory closer to processor. But memory access and computation remain as two separate steps, which does not address the cost of data communication and memory access, which impose the most overhead in both neuromorphic and general purpose computing systems.


The memristor has been presented as an option to merging the computation and memory substrates \cite{chua1971memristor, chua1976memristive, strukov2008missing}. What was once a theoretical postulate proposed by Chua in 1971 \cite{chua1971memristor} has become a commercially available technology that can be integrated in the back-end-of-the-line (BEOL) of modern CMOS processes \cite{eflash, Mad200}. Their non-volatile retention capacity is often likened to synapses \cite{jo2010nanoscale, rahimi2020complementary, cai2019fully, serrano2013stdp, eshraghian2022memristor}, and their threshold-switching characteristics occasionally exploited as a spiking neuron model \cite{pickett2013scalable, eshraghian2018neuromorphic, zhang2017artificial, zhang2020brain}. For example, the modulation of device resistance has been correlated to synaptic plasticity, where achieving short-term plasticity (STP) and long-term plasticity (LTP) using memristors is as simple as applying programming pulse trains \cite{wang2020integration}. The benefits of memristive neurons and synapses arise from CMOS-compatibility, high-density, nanoscale vertical integration, and their ability to directly implement biological features, as opposed to requiring several arithmetic steps as with transistor-only circuits \cite{covi2016analog}. The use of spiking neural networks (SNNs) as opposed to modern deep learning paradigms has shown significant energy and latency benefits as a result of activation sparsity, spike-based representations of data, and event-based data processing \cite{eshraghian2021training, azghadi2020hardware, eshraghian2022fine, zhou2022spiceprop, zhou2020towards}, and can be harnessed using the growing infrastructure to support SNN simulations \cite{stimberg2019brian, gewaltig2007nest, bekolay2014nengo, davison2009pynn, hines2001neuron}.

Much of the prior work on memristive spiking neural networks (MSNNs) is constrained to using either memristive synapses \textit{or} memristive neurons. The work in \cite{wang2018fully} demonstrated a fully integrated memristive system that emulated both synapses and spiking neuron models to achieve simple pattern recognition tasks. The result is demonstrated using an in-house fabricated diffusive memristor, which is not readily accessible to the broader research community. In this work, we introduce a fully memristive approach to designing a system that shows the capacity to learn pattern recognition tasks using memristive neurons and synapses. We achieve this using models of commercially available, low-cost memristors \cite{molter2016generalized}. We use a memristive neuron circuit which consists of the minimum number of circuit elements necessary to achieve distinct depolarization, hyperpolarization, and repolarization voltage phases. This neuron is integrated in a multi-layer MSNN with memristive synaptic interconnections, modelled using an equivalent SPICE model of a memristive array to train the MSNN with the STDP learning rule \cite{bi1998synaptic}. We successfully demonstrate the potential of MSNN to perform associative memory retrieval, as well as pattern recognition achieving 97.5\% accuracy on a multi-class classification task. Our results validate homogeneous memristive systems that are capable of neuromorphic learning without global error signals.


\section{Methods}

\subsection{MIF model}

In this paper, we apply the MIF neuron model \cite{kang2021build}. The MIF circuit is shown in Fig.~\ref{mif2circuit}. The proposed MIF circuit is characterized by the differential equations below:

\begin{figure}[!htbp]
\includegraphics[scale=0.28]{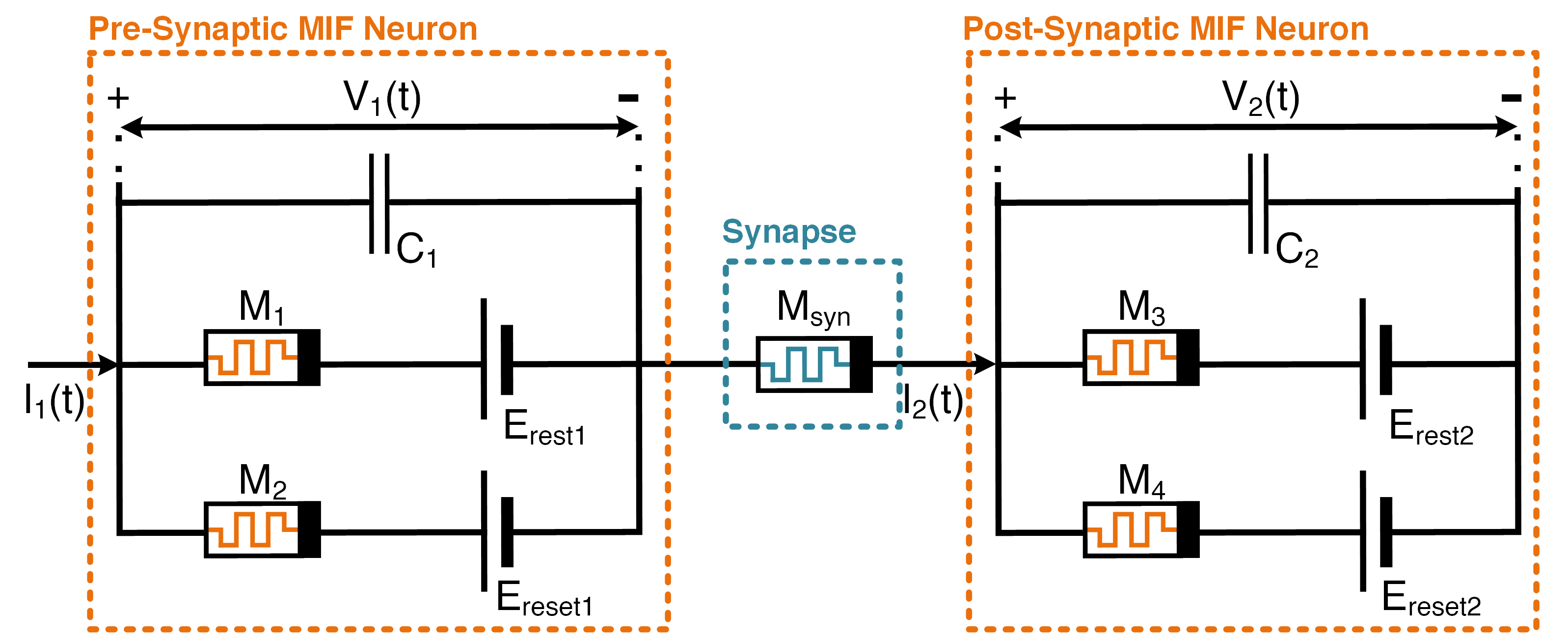}
\caption{Pre-synaptic and post-synaptic memristive neurons with a trainable memristive synapse interposed between the two. The neuron model is a memristive integrate-and-fire (MIF) neuron circuit, consisting of two memristors $M_1$ and $M_2$, connected to DC voltage sources $E_{\rm rest}$ and $E_{\rm reset}$, in parallel with a capacitor $C$. Voltage spikes generated by the MIF neuron propagates through the synapse, and trigger an input current to the post-synaptic MIF neuron, which in turn will generate spikes.}
\vspace{-16pt}
\label{mif2circuit}
\end{figure}

\begin{subequations}\label{mif2}
\begin{equation}
\frac{{\mathrm{d}}v}{\mathrm{d}t}=\frac{I - G_1(v-E_{rest}) - G_2(v-E_{reset})}{C} 
\end{equation}
\begin{equation}
\frac{{\mathrm{d}x_1}}{\mathrm{d}t}= \frac{1}{\tau_1}(\frac{1-x_1}{1+e^{\frac{v_{on1}-(v-E_{rest})}{k_{th}}}}  -   \frac{x_1}{1+e^{\frac{(v-E_{rest})-v_{off1}}{k_{th}}}}) 
\end{equation}
\begin{equation}
\frac{{\mathrm{d}x_2}}{\mathrm{d}t}= \frac{1}{\tau_2}(\frac{1-x_2}{1+e^{\frac{v_{on2}-(v-E_{rest})}{k_{th}}}}  -   \frac{x_2}{1+e^{\frac{(v-E_{rest})-v_{off2}}{k_{th}}}}) 
\end{equation}
\begin{equation}
G_1 = \frac{x_1}{R_{on1}} + \frac{1-x_1}{R_{off1}} 
\end{equation}
\begin{equation}
G_2 = \frac{x_1}{R_{on2}} + \frac{1-x_2}{R_{off2}} 
\end{equation}
\end{subequations}

\noindent where $G_1$ and $G_2$ are the memductances, $x_1$ and $x_2$ are a pair of internal states, $\tau_1$ and $\tau_2$ are time constants governing the rate of change in internal states, $k_{th}$ is the effective thermal voltage. They are the characteristic variables of $M_1$ and $M_2$, respectively. This system of equations mirrors several prominent SPICE memristor models, and has been used to emulate the commercially available Knowm memristor \cite{molter2016generalized}.

To show the dynamics of the above system, we apply an alpha-shaped input current, with the resultant memristor internal states and voltage waveforms shown below in Fig.~\ref{mif2result}:

\begin{figure}[htbp]
  \begin{center}
  \includegraphics[width=3in]{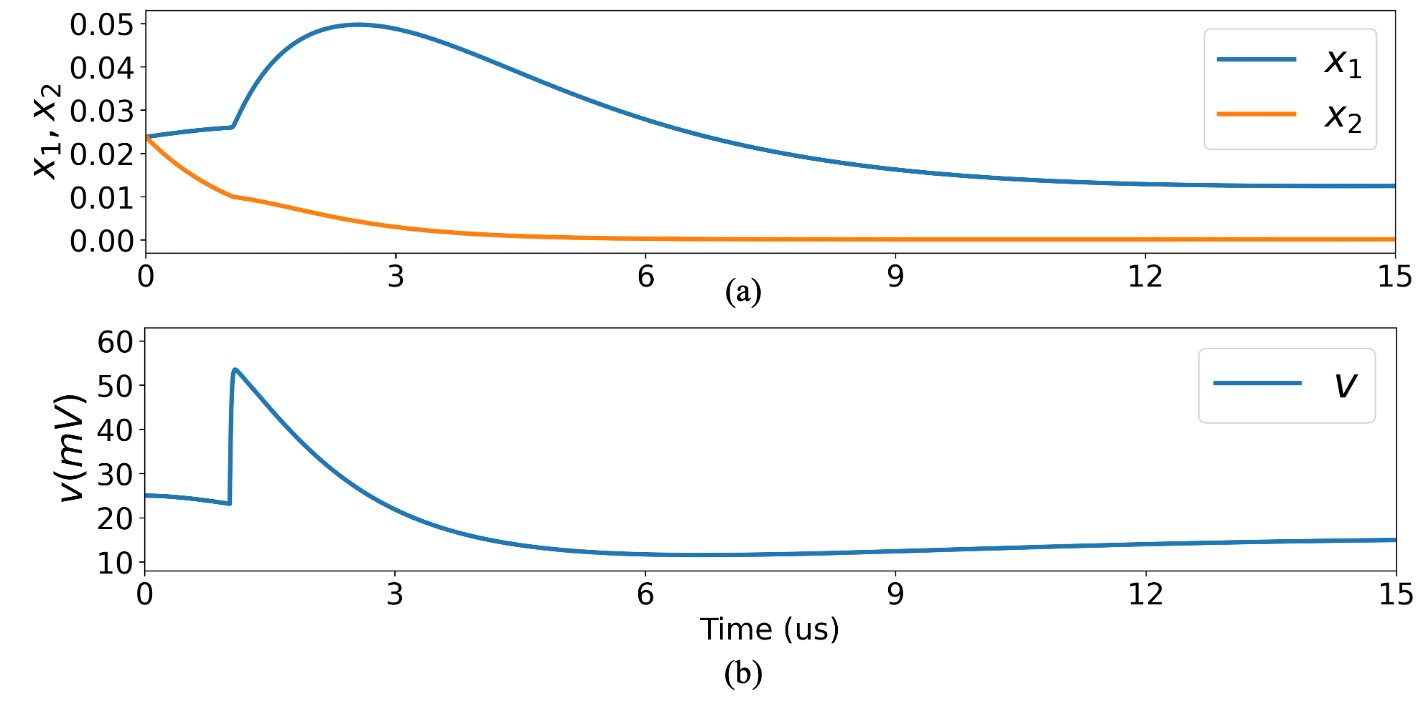}
  \vspace{-8pt}
  \caption{Simulation result of MIF receiving an alpha shape input current. \\
  (a) Internal states $x_1$, $x_2$. (b) Voltage response $v$. }\label{mif2result}
  \end{center}
\end{figure}

\vspace{-8pt}
The alpha current is modeled by Eq.~(2):
\begin{subequations}\label{alphaCurrent}
\begin{align}
\tau_{syn} \frac{{\mathrm{d}I}}{\mathrm{d}t} &= a-I \\
\tau_{syn} \frac{{\mathrm{d}a}}{\mathrm{d}t} &= -a + W_j \cdot \sum_f{\delta(t-t_j^f)}
\end{align}
\end{subequations}

\noindent where $W_j$ is a weighted synaptic current generated between presynaptic neuron $j$ and associated postsynaptic neuron. $\sum_f{\delta(t-t_j^f)}$ indicates the total number of spikes emitted by presynaptic neuron $j$, and $I$ is the input current.

The parameters used in this network are listed in Table \ref{tablemsnn1}:
\vspace{-6pt}
\begin{table}[htb]
\caption{MIF circuit parameters}
\vspace{-10pt}
\label{tablemsnn1}
\begin{center}
\begin{tabular}{|c|c|c|c|}
\hline
\textbf{Parameter} & \textbf{Value} & \textbf{Parameter} & \textbf{Value} \\ \hline 
$E_{rest}$ & 0 mV & $E_{reset}$ & 50 mV \\
$C$ & 100 pF & $k_{th}$ & 15 mV \\
$v_{off_1}$, $v_{off_2}$ & 0 mV & $v_{on_1}$, $v_{on_2}$ & 100 mV \\
$R_{off_1}$, $R_{off_2}$ & 0.1 M$\Omega$ & $R_{on_1}$, $R_{on_2}$ & 1 k$\Omega$ \\
$\tau_1$, $\tau_2$ & 1 $\mu$s & & \\
\hline
\end{tabular}
\end{center}
\footnotesize{}
\vspace{-10pt}
\end{table}

\vspace{-8pt}
\subsection{Memristive Synapse with STDP}
The Spike-timing-dependent plasticity (STDP) learning rule modulates synaptic weights based on the time difference of pre- and postsynaptic spike arrivals \cite{bi1998synaptic}. This type of learning rule can be be achieved by using memristors \cite{maranhao2021low, serrano2013stdp}, and can be verified using SPICE-level models \cite{yakopcic2013generalized}. As Fig.~\ref{mif2circuit} shows, a synaptic memristor is interposed between two neurons, where the pre- and postsynaptic spikes will generate a voltage across the memristor that causes the memristance to be updated. Moreover, the time difference between pre- and postsynaptic spikes will modulate the changes in the memristance of the synapse.


In the memristive synapse with STDP learning mechanism, the weight change will increase rapidly and then decrease slowly when the time difference of a pre- and post- synaptic spike increases from zero. The weight change of a memristive synapse is determined by the voltage across the memristor. The largest voltage will result in the largest weight change, and occurs when one of the pre- or post- synaptic neuron reaches the threshold level, while the other is at the reset voltage level. Therefore, the learning window of memristive STDP will rise with a fast time constant, followed by a slow decay. Thus, it is reasonable to model the memristive synapse behavior with an alpha shape learning window as:

\vspace{-8pt}
\begin{subequations}\label{stdpzp}
\begin{align}
W_{pre}(x) &= U_{pre} \cdot \frac{\abs{\Delta t}}{\tau_{pre}} \cdot e^{-\frac{\abs{\Delta t}}{\tau_{pre}}} \quad \\
\text{at} & \quad t_{post} \quad \text{for}\quad t_{pre} < t_{post} \nonumber \\
W_{post}(x) &= U_{post} \cdot \frac{\abs{\Delta t}}{\tau_{post}} \cdot e^{-\frac{\abs{\Delta t}}{\tau_{post}}} \quad \\
\text{at} & \quad t_{pre} \quad \text{for}\quad t_{post} < t_{pre} \nonumber
\end{align}
\end{subequations}

\noindent where U$_{\rm pre}$, U$_{\rm post}$, $\tau_{\rm pre}$, $\tau_{\rm post}$ are fitted constants, and $\Delta$t=t$_{\rm post}$-t$_{\rm pre}$. Generally, the curve in the learning window will take an alpha shape in both the positive and negative planes of the $x$-axis. 

This type of STDP can be modeled with a system of differential equations defined in Eq.~(4):

\begin{subequations}\label{mstdpEq}
\begin{equation}
\tau_{pre} \frac{{\mathrm {d}} A_{pre}}{\mathrm {d} t}=T_{pre}-A_{pre}
\end{equation}
\begin{equation}
\tau_{pre} \frac{{\mathrm {d}} T_{pre}}{\mathrm {d} t}=T_{pre} + U_{pre} \cdot \sum_f{\delta(t-t_j^f)}
\end{equation}
\begin{equation}
\tau_{post} \frac{{\mathrm {d}} A_{post}}{\mathrm {d} t}=T_{post}-A_{post}
\end{equation}
\begin{equation}
\tau_{post} \frac{{\mathrm {d}} T_{post}}{\mathrm {d} t}=-T_{post} + U_{post} \cdot \sum_n{\delta(t-t_i^n)}
\end{equation}
\begin{equation}
W_j \leftarrow W_j + A_{post}\quad \text{upon presynaptic spike}
\end{equation}
\begin{equation}
W_j \leftarrow W_j + A_{pre}\quad \text{upon postsynaptic spike}
\end{equation}
\end{subequations}

\noindent where $f$ indicates the total numbers of spikes emitted by the presynaptic neuron $j$, and $n$ defines the total number of spikes emitted by the postsynaptic neuron $i$. U$_{\rm pre}$ is typically positive, while U$_{\rm post}$ is usually negative. $W_j$ is the weighted synaptic current. Upon arrival of a presynaptic spike, $W_j$ immediately increases by the amount of A$_{\rm post}$. T$_{\rm pre}$ induces a rise by the amount of U$_{\rm pre}$, which then evolves according to Eq.~(4b). This in turn modulates A$_{\rm pre}$ (4a). Upon arrival of a postsynaptic spike, $W_j$ immediately increases by the amount of A$_{\rm pre}$. T$_{\rm post}$ increases by the amount of U$_{\rm post}$, and then effects A$_{\rm post}$ in a similar manner to the impact from a presynaptic spike.

\begin{figure}[htbp]
  \begin{center}
  \includegraphics[width=3.4in]{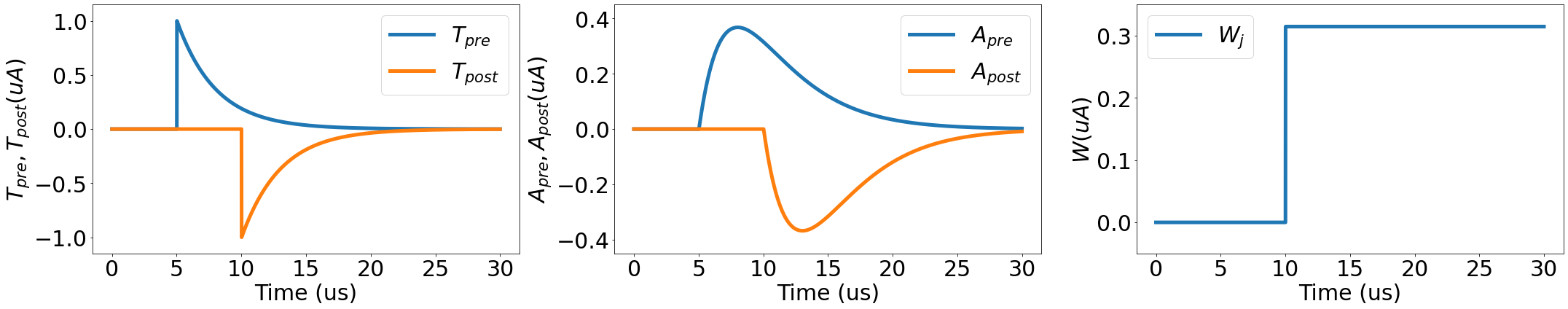}\\
  \caption{An example of the proposed memristive STDP with $\tau_{\rm pre}$ = $\tau_{\rm post}$=3 $\mu s$, U$_{\rm pre}$=1 $\mu A$, U$_{\rm post}$=-1 $\mu A$. (a) T$_{\rm pre}$ and T$_{\rm post}$ in Eq.~(4) are determined by a pre- and a post- synaptic spike, respectively. (b) A$_{\rm pre}$ and A$_{\rm post}$ in Eq.~(4) are determined by a pre- and a post- synaptic spike, respectively. (c) The weight is updated according to Eq.~(4).}\label{mstdpFig}
  \end{center}
\end{figure}

An example of the temporal evolution of the learning rule is shown in Fig.~\ref{mstdpFig}, where an update occurs if presynaptic spike arrival occurs before the postsynaptic spike. Fig.~\ref{mstdpFig}(a) shows T$_{\rm pre}$ and T$_{\rm post}$ instants, at which the curve takes an immediate increase followed by an exponential decay. Fig.~\ref{mstdpFig}(b) shows A$_{\rm pre}$ and A$_{\rm post}$ as functions of time, which have alpha-shaped curves, and Fig.~\ref{mstdpFig}(c) shows the fast weight increase when the presynaptic spike occurs before the postsynaptic spike. When the presynaptic spike arrives at $t=5 ms$, $W_j$ increases by the amount of A$_{\rm post}$ at this time, which is zero. At $t = 10 ms$, the postsynaptic spike arrives, $W_j$ increases by the amount of A$_{\rm pre}$ at the this time, which is around 0.3 $\mu A$.  

\vspace{-4pt}





\vspace{-16pt}
\subsection{Type-1 MSNN for Memory Retrieval}
\begin{figure}[htbp]
  \begin{center}
  \includegraphics[width=2.8in]{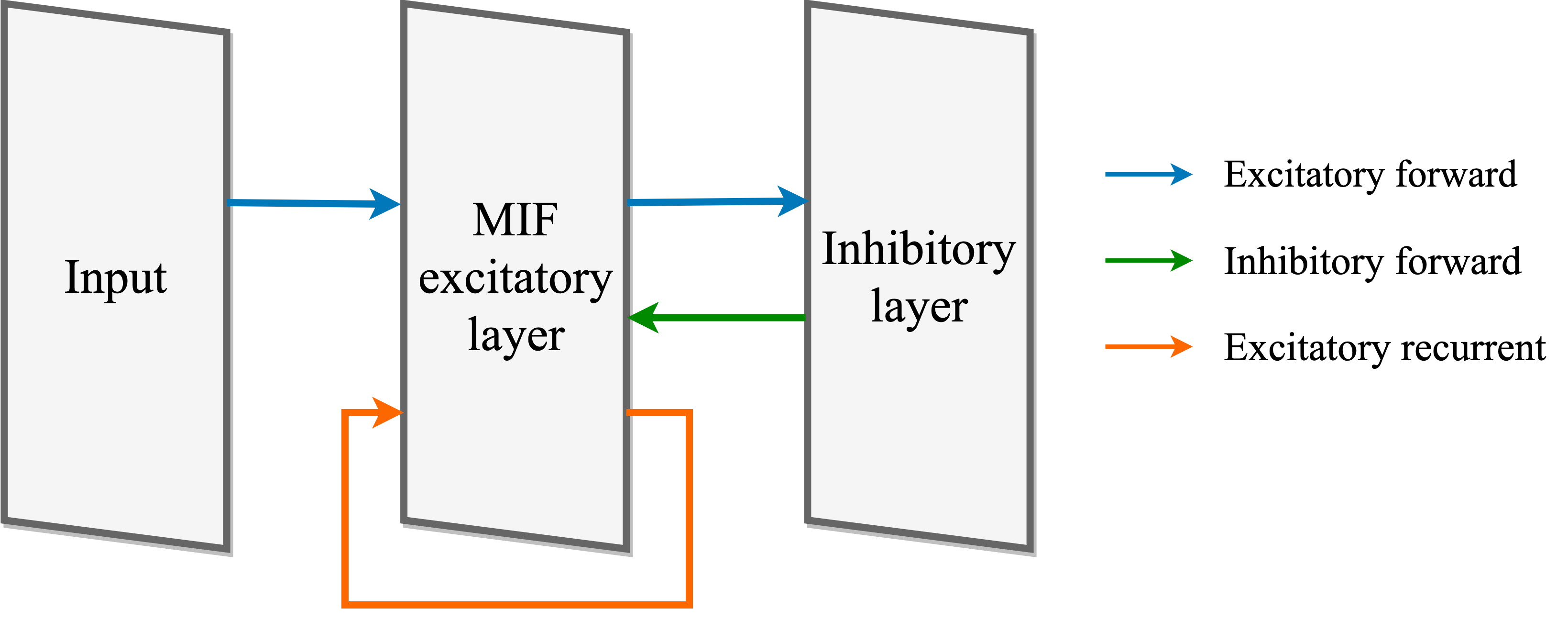}\\
  \caption{The memristive spiking neural network architecture, which consists of the input layer, MIF excitatory layer, and the inhibitory layer. The blue arrows denote the excitatory synaptic forward connections, the orange arrow indicates excitatory synaptic recurrent connections, and the green arrow shows the inhibitory synaptic forward connection.}\label{mifsnn1}
  \end{center}
\end{figure}

The neural network architecture we implement with the fully memristive neuron and synapse models consists of three layers \cite{diederich2018memristive}. The input layer applies a Poisson spike train which encodes the input patterns via rate encoding. The second layer is a MIF excitatory layer. The third layer is an inhibitory layer and includes feedback connections to the excitatory layer. The architecture is shown in Fig.~\ref{mifsnn1}. The input to the excitatory layer consists of excitatory synapses with fixed weights between the input layer and the MIF excitatory layer, with one-to-one connections. The output of the excitatory layer includes trainable (via STDP) excitatory recurrent synaptic connections, in addition to fixed-weight excitatory synapses between the MIF excitatory layer and the inhibitory layer, also with one-to-one connections. The inhibitory layer includes inhibitory feedback synapses with fixed weights between the inhibitory layer and the MIF excitatory layer with one-to-(all-1) connections (not including the single MIF neuron connected to the inhibitory neuron). In this neural network, the number of neurons in all three layers are identical, and STDP is enabled within the MIF excitatory layer. The input patterns consist of 32$\times$32 pixels, such that the total number of neurons in each layer is 1024. Each pattern generates a Poisson spike train of 35 $\mu s$ duration, followed by a 15 $\mu s$ refractory period without any input such that each input pattern does not affect the network dynamics upon arrival of the next input pattern. Each input spike will generate an alpha current where $\tau_{\rm syn}=10 ns$. The threshold of each MIF neuron is set to 25 $mV$. As in the MIF neuron, there is no extra control circuit to force reset, and we set a refractory of 3 $\mu s$ to prevent duplicating spikes during the simulation. We find the weight increase has a more significant effect when compared with the weight decrease, and it also enables the network to learn faster. Therefore, we set U$_{\rm pre} = 10 \mu A$ and U$_{\rm post} = -0.1 \mu A$. Note that U$_{\rm pre,post}$ refers to an internal state variable that is modulated by the current-based neuron model, and is in dimensions of amperes. The recurrent connections are initialized with randomly weighted synaptic currents between 0 to 0.2 $\mu A$. The exact weight values between excitatory layer and inhibition layer do not have a large effect, which enables the inhibitory neurons to trigger a spike following an output from excitatory neurons. On the other hand, the weights between input-to-excitatory layers and the inhibitory feedback mechanism both need to be carefully tuned to be neither too weak, nor too strong, to prevent complete suppression of downstream spikes, or excessive reinforcement of firing. We chose the fixed input to the excitatory weight to be 50 $\mu A$, and the fixed inhibitory layer feedback weight to be 20 $\mu A$.

\vspace{-16pt}
\subsection{Type-2 MSNN for Pattern Recognition}
The sequence of the MSNN for pattern recognition is identical to the memory retrieval network, in that we use a Poisson input layer, an excitatory layer, and an inhibitory layer. To improve unsupervised learning we modify the nature of the synaptic connections. STDP-based learning is enabled between the input layer and the excitatory MIF layer with all-to-all connections. As before, each excitatory neuron connects to one inhibitory neuron and the inhibitory layer uses inhibitory synapses with fixed weights connected to the excitatory layer. In this network, no recurrent connections are present due to the lack of temporal dependencies in the pattern recognition task. The number of neurons between the excitatory and inhibitory layers must be identical, though they no longer need to match the input layer. We use 1024 neurons in the input layer, and 320 neurons each in the excitatory and inhibitory layers, where STDP takes place between the input and the excitatory layers. We set U$_{\rm pre} = 10 nA$, U$_{\rm post} = -0.1 nA$, and the fixed inhibitory layer feedback weight to 200 $\mu A$. The weighted synaptic current between input and excitatory layers are randomly initialization between 0 and 12 $\mu A$ based on empirical evaluation. Additionally, a random 0-2 $\mu s$ delay between input layer and excitatory layer is assigned to avoid excessive simultaneous spiking, thus mitigating the exploding weights problem. All other neuron and input parameters are identical to the memory retrieval tasks.


\vspace{-4pt}
\section{Results}
\subsection{Type-1 MSNN for Memory Retrieval Result}
\begin{figure}[htbp]
  \begin{center}
  \vspace{-8pt}
  \includegraphics[width=2.8in]{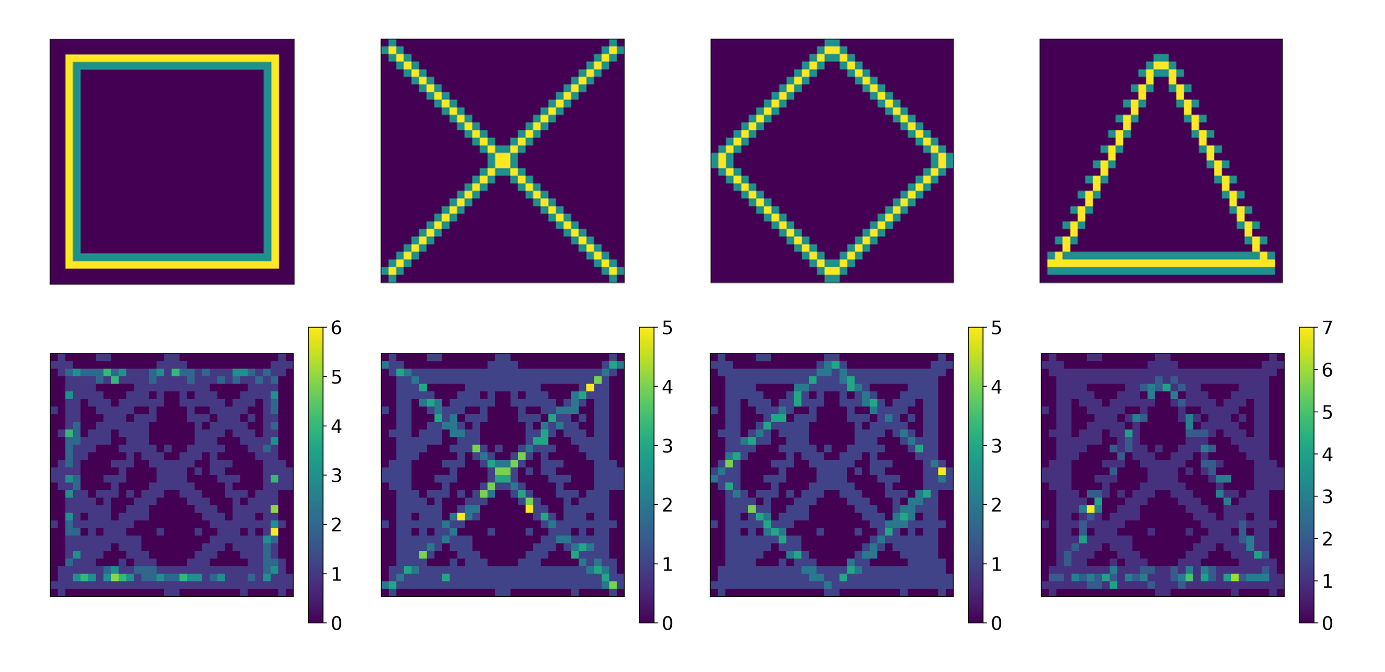}\\
  \vspace{-8pt}
  \caption{Top row: the four input spiking patterns (1:Square, 2: Cross, 3: Diamond, 4: Triangle) applied to the input of the MSNN. Bottom row: the number of spikes in the excitatory layer during the test stage, upon receiving these four input patterns.}\label{inputPattern}
  \end{center}
\end{figure}

We provided four different patterns to the MSNN successively for each iteration, and simulated across multiple epochs. This is to coarsely emulate how biological systems experience real-world data in a batch size of `1' in online learning systems. The Poisson input layer converts each pixel intensity of the input pattern into a spiking probability of each neuron. We used four different patterns, as shown in the top row of Fig.~\ref{inputPattern}. The weight matrix is updated with the STDP rule. The training phase consists of 20 epochs for all patterns, corresponding to 80 iterations.


\begin{figure}[htbp]
  \begin{center}
  \vspace{-8pt}
  \includegraphics[scale=.28]{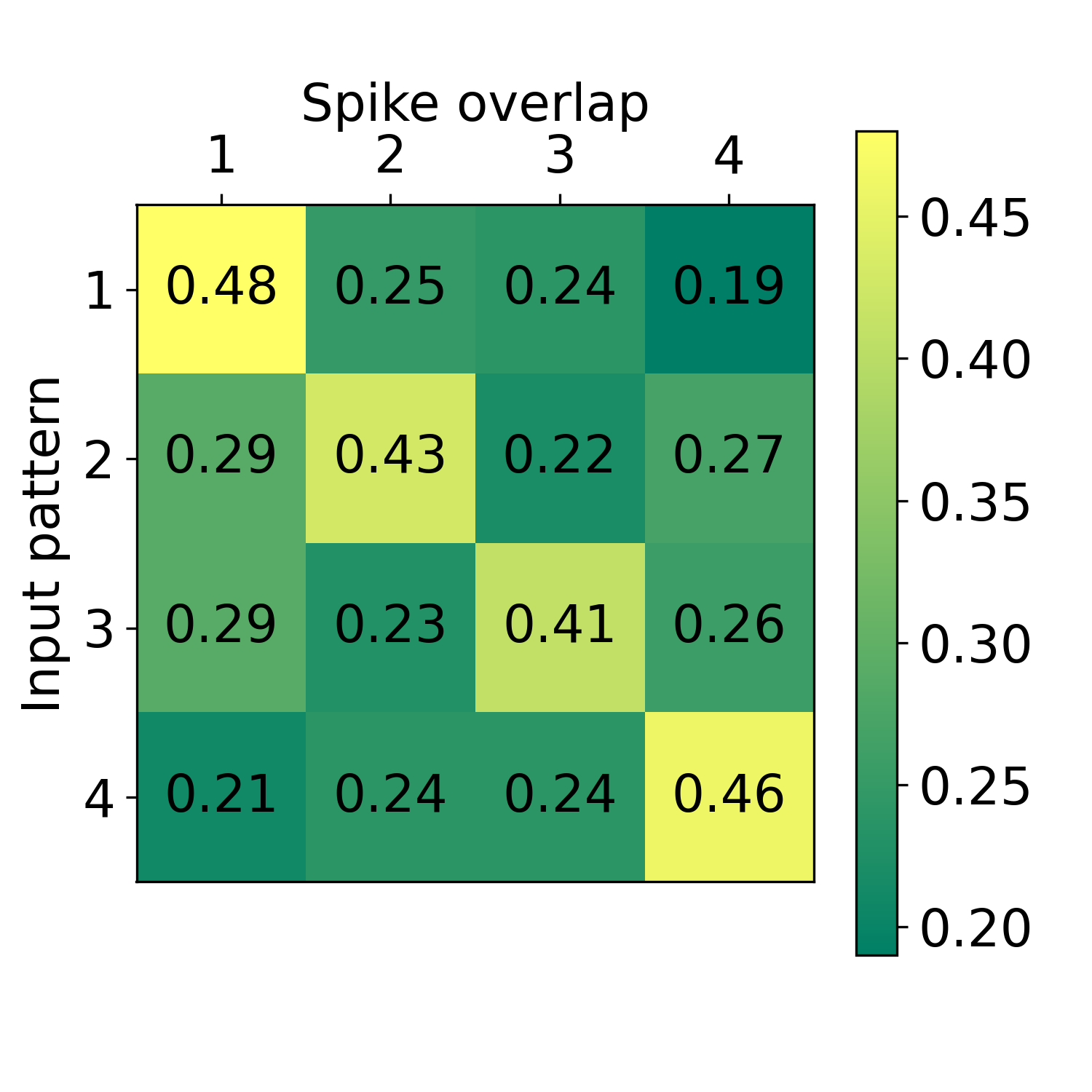}
  \vspace{-16pt}
  \caption{The resultant heatmap shows the memory retrieval resemblance for four patterns, which are 1:Square, 2: Cross, 3: Diamond, 4: Triangle.}\label{matrix}
  \end{center}
  \vspace{-8pt}
\end{figure}

After training, the same four patterns are successively applied to the MIF. The MSNN is expected to `recall' the patterns as the connections have ideally made associations in the form of synaptic memory. When one of the patterns is applied to the MSNN again, other patterns are also recalled with less intensity, which is related to `memory retrieval', illustrated in the bottom row of Fig.~\ref{inputPattern}. In order to evaluate memory retrieval performance, we calculate the percentage of overlapping spike count with incorrect patterns, with the results shown in Fig.~\ref{matrix}. We find that when an input pattern is provided to the network, its resemblance to the spiking behavior with the target behavior is at a maximum. This is depicted in Fig.~\ref{matrix}, where the values along the diagonal are the largest. 
As shown in the bottom row of Fig.~\ref{inputPattern}, for each input pattern, the network continues to recall features of the other three patterns, which is a result of low-resistance pathways that cause overlap with other patterns, but with less resemblance to the target pattern. 

\vspace{-8pt}

\subsection{Type-2 MSNN for Pattern Recognition Result}
\begin{figure}[htbp]
  \begin{center}
  \includegraphics[width=2.8in]{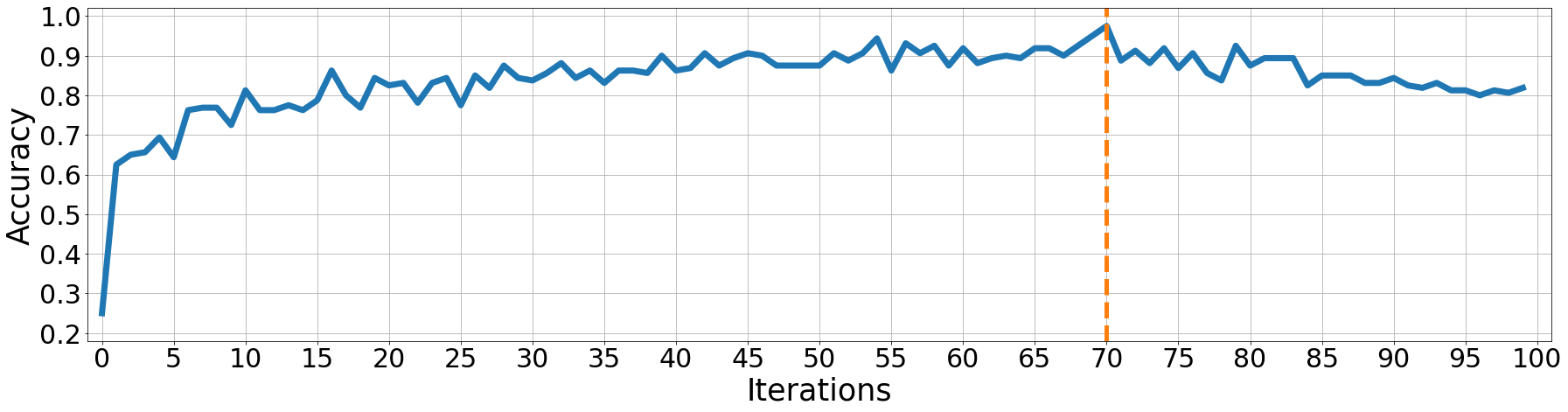}
  \caption{Accuracy across multiple iterations.}\label{accuracyResult}
  \end{center}
  \vspace{-16pt}
\end{figure}

During training, we applied the four different patterns (Fig.~\ref{inputPattern}) to the MSNN successively, then accumulated and recorded the number of spikes in each output MIF neuron for each pattern. During the training process, each MIF neuron is randomly assigned a pattern and updated after every 5 epochs. This allows each MIF neuron to be associated with a pattern such that it may be used to encode one of four patterns during the test stage. After assignments are updated, we use the current weights to test the accuracy of the network. During the test stage, the four patterns are passed to the network for 40 epochs and the total number of spikes are counted for each pattern for all output excitatory neurons. The neuron with the highest spike count is deemed as the winner neuron for this input pattern. The accuracy evolution during training is shown in Fig.~\ref{accuracyResult}, where a total accuracy of $97.5\%$ is attained at the 70th iteration, shown by the dashed orange line. Early stopping is applied here, as further training causes excessive reinforcement of high-conductance pathways. 




\vspace{-4pt}
\section{Discussion and Conclusion}
In this paper, we proposed two different types of fully memristive neural networks for unsupervised learning, based on a memristive integrate-and-fire neuron model together with memristive synapses. The synapses and neuron models are designed using SPICE-level memristor models to relate circuit-level plausibility with the biological plausibility of spiking neurons. 
We demonstrated memory retrieval and pattern recognition across the four input patterns, and validated the potential of our fully-memristive approach across both tasks.
Future work will address more challenging tasks, such as testing memory retrieval for incomplete patterns, and increasing dataset complexity for the multi-pattern classification problems. Explorations of new learning paradigms that rely on error-propagation may need to be integrated together with the STDP update rule to enable the success of more complex tasks. 
The proposed framework will render circuit-level implementations using a broad class of memristors to perform neuromorphic computing.



%


\ifCLASSOPTIONcaptionsoff
  \newpage
\fi

\bibliographystyle{IEEEtran}
\bibliography{references}

%






\end{document}